%% file: 3d_pose.tex
\documentclass[journal]{IEEEtran}

\usepackage{amsmath}
\usepackage{amssymb}
\usepackage{multirow}
\usepackage{dsfont}
\usepackage{comment}
\usepackage{booktabs}
\usepackage{graphicx}
\usepackage{times}
\usepackage{epsfig}
\usepackage{color}

\newcommand{\eg}{\textit{e.g.}{}{\@}}
\newcommand{\etal}{\textit{et al.}{}{\@}}
\newcommand{\ra}{$ \rightarrow ${}{}\@}

\ifCLASSINFOpdf
\else
\fi

\begin{document}

\title{View Invariant 3D Human Pose Estimation}

\author{Guoqiang Wei, Cuiling Lan,~\IEEEmembership{Member,~IEEE}, Wenjun Zeng,~~\IEEEmembership{Fellow,~IEEE,} and Zhibo Chen,~\IEEEmembership{Senior Member,~IEEE}
	\thanks{
		This work was done while Guoqiang Wei was an intern with Microsoft Research Asia. 
		
		\vspace{1.0em}
		Guoqiang Wei and Zhibo Chen are with the Department of Electronic Engineering and Information Science, University of Science and Technology of China. (e-mail: wgq7441@mail.ustc.edu.cn; chenzhibo@ustc.edu.cn)
		
		\vspace{1.0em}
		
		Cuiling Lan and Wenjun Zeng are with Microsoft Research Asia. (e-mail: culan@microsoft.com; wezeng@microsoft.com)}}

\maketitle

\begin{abstract}
The recent success of deep networks has significantly advanced 3D human pose estimation from 2D images. The diversity of capturing viewpoints and the flexibility of the human poses, however, remain some significant challenges. In this paper, we propose a view invariant 3D human pose estimation module to alleviate the effects of viewpoint diversity. The framework consists of a base network, which provides an initial estimation of a 3D pose, a view-invariant hierarchical correction network (VI-HC) on top of that to learn the 3D pose refinement under consistent views, and a view-invariant discriminative network (VID) to enforce high-level constraints over body configurations. In VI-HC, the initial 3D pose inputs are automatically transformed to consistent views for further refinements at the global body and local body parts level, respectively. For the VID, under consistent viewpoints, we use adversarial learning to differentiate between estimated poses and real poses to avoid implausible 3D poses.  
Experimental results demonstrate that the consistent viewpoints can dramatically  enhance the performance. Our module shows robustness for different 3D pose base networks and achieves a significant improvement (about 9\%) over a powerful baseline on the public 3D pose estimation benchmark Human3.6M.	
\end{abstract}


%
\IEEEpeerreviewmaketitle

\input{intro}
\input{related_work}
\input{method}
\input{experiments}

\input{conclusion}

\ifCLASSOPTIONcaptionsoff
  \newpage
\fi

\bibliographystyle{IEEEtran}
\bibliography{IEEEabrv,3d_pose}

\end{document}

%% file: intro.tex
\section{Introduction}

Estimating 3D human pose from an RGB image has been an active research field for many years. It facilitates a wide spectrum of applications such as human computer interaction, action recognition, sports performance analysis, and augmented reality \cite{sarafianos20163d}. This is still a challenging task as it should not only overcome the barriers that exist in 2D pose estimation such as the diversities in viewpoint, clothing, lighting and the flexibility in body articulation, but also resolve the ambiguities in recovering depth from a 2D projection of 3D objects. The development of neural networks has advanced the 3D human pose estimation \cite{zhou2017towards,sun2017compositional,pavlakos2017volumetric,moreno20173d,martinez2017a,fang2018learning}. Generally, previous methods are categorized into two classes: i) training an end-to-end network to directly predict 3D pose from an image \cite{zhou2017towards,sun2017compositional,pavlakos2017volumetric}, ii) estimating 2D pose from an image first and then lifting the 2D pose to 3D pose \cite{moreno20173d,chen20173d,martinez2017a,fang2018learning}.

Despite the general success of the end-to-end learning paradigm, two-step solutions which consist of a convolutional neural network for predicting 2D joint locations from an image and a subsequent optimization step to recover the 3D pose also win excellent performance \cite{pavlakos2017volumetric,martinez2017a,fang2018learning}. One reason is that 2D pose estimation \cite{newell2016stacked,wei2016cpm,chou2017self,chen2017adversarial} has gained significant advances and can provide plausible estimation with high accuracy \cite{chen20173d}. To infer a 3D pose from 2D pose input, many approaches have been designed, \eg, by memorization ({\it{i.e.}}, matching) \cite{jiang20103d,yasin2016a,chen20173d} or by regressing the 3D pose \cite{moreno20173d,martinez2017a,fang2018learning}. Even though promising results have been achieved, few approaches explore the challenge caused by the diversity of viewpoints. In practical scenarios, a person can be captured from an arbitrary viewpoint by a camera. Similarly, a person can perform an action towards different orientations with respect to a camera. These result in diverse viewpoints for both the human poses. It is generally challenging for one model to tackle poses with diverse viewpoints.  

\begin{figure*}[pt]
	\includegraphics[width=1.0\textwidth]{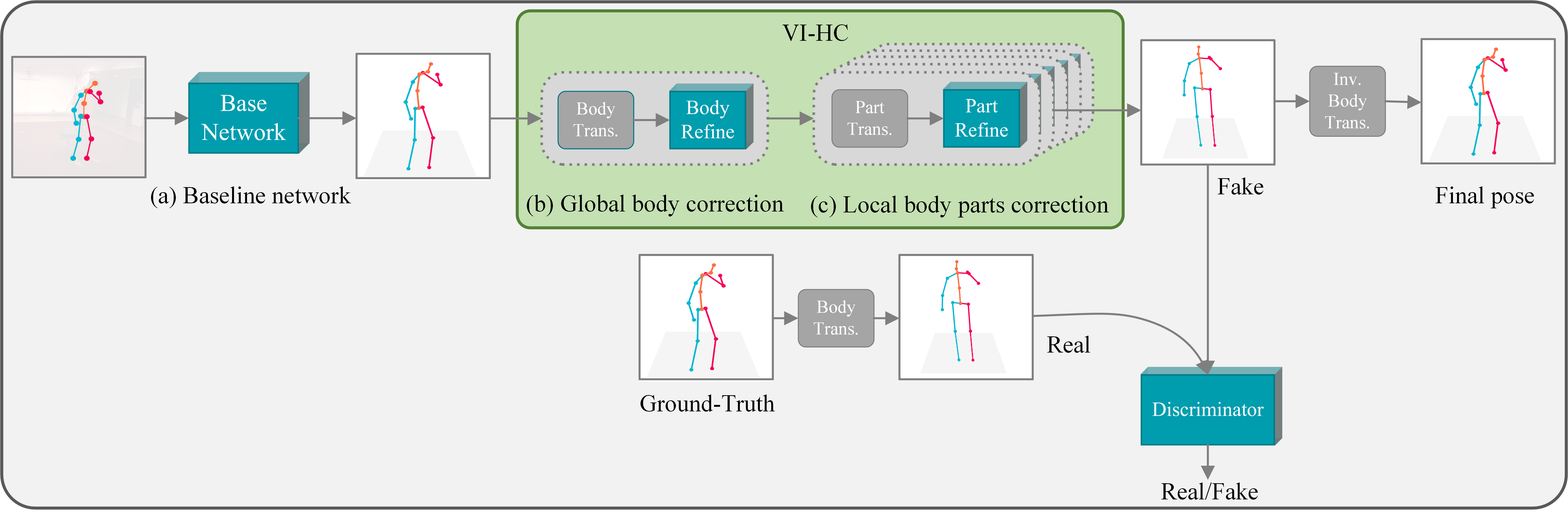}
	\caption{
		The proposed framework for view invariant (VI) 3D pose estimation. Given a set of 2D joint locations of a person, the base network (a) predicts an initial 3D pose $ \mathbf{\hat{S}}_{init}$, which is then further corrected by the proposed (b) global view-invariant body correction and the (c) local view-invariant body parts correction subnetworks (note that the Inv. Part Trans. module following \emph{Part Refine} are not shown to save space) to generate refined 3D poses. Moreover, we enforce high-level body configuration constraint during the training by adversary learning, where a view invariant discriminator is jointly trained with the generator (\eg, our 3D pose estimator) to distinguish the ground-truth poses from the generated ones under consistent views. Through Inv. Body Trans., the 3D pose is finally inversely transformed back to the original view as the final output.
	}
	\label{fig:pipeline}
\end{figure*}

In this work, we propose a general view invariant module to refine 3D poses generated from base networks. It alleviates the influence of viewpoint diversity by automatically transforming the intermediate 3D poses to a consistent view, facilitating 3D pose correction. As shown in Figure \ref{fig:pipeline}, we build our framework by stacking a base 3D pose estimation network, and a view-invariant hierarchical correction network (VI-HC), to infer the 3D poses. To explore the global structure of the human body and the flexible configurations of local parts, VI-HC is constructed by a global view-invariant correction subnetwork and a local body parts view-invariant correction subnetwork, which efficiently refines the 3D poses by removing the influence of diverse viewpoints and flexibility of body parts. Moreover, to enforce high-level constraints over body configurations, a simple discriminator is added \emph{under the consistent views} as a high level loss for training, which efficiently distinguishes the plausible 3D poses from fake 3D poses. Note that the consistent view easies both the refinements from VI-HC and discriminative correction from the discriminator. We validate the effectiveness of the view-invariant module on top of two different powerful base 3D pose estimation networks respectively. Experimental results demonstrate our proposed module improves the performance by about 9\% and 3\% on the two base networks respectively, on the public 3D human pose estimation benchmark Human3.6M dataset. 



In summary, we make four main contributions:

\begin{itemize}
	\setlength{\itemsep}{0pt}
	\item We design a simple but powerful view invariant scheme to address the challenge of diverse viewpoints to enhance the 3D pose estimation performance.
	\item Hierarchical view invariant correction subnetworks are designed to refine the 3D poses under consistent views with respect to the global body, and the five articulated body parts, respectively. 
	\item We propose the use of a simple view invariant discriminator to enforce a high-level constraint over body configurations to exclude implausible 3D poses.
	\item Our module consistently improves the performance and shows robustness over different baselines. 
\end{itemize}

%% file: related_work.tex
\section{Related Work}
\subsection{3D Pose Estimation}

Over the recent years, we have witnessed the tremendous progress in the field of 3D pose estimation. Tekin \etal \cite{tekin2016direct} rely on an auto-encoder to learn high-dimensional latent pose representation and regress 3D poses from 2D images. Pavlakos \etal  \cite{pavlakos2017volumetric} train a convolutional neural network to predict per voxel likelihoods for each joint in a fine discretized 3D volumetric representation. Sun \etal  \cite{sun2017compositional} propose to regress the joint locations and bone representations to exploit structure information. Extra 2D pose datasets are usually utilized to enhance the performance \cite{mehta2017monocular}.

Different from those end-to-end 2D image to 3D pose regression approaches \cite{tekin2016direct,pavlakos2017volumetric,sun2017compositional}, Martinez \etal \cite{martinez2017a} show that a well-designed simple network for regressing 3D pose from 2D pose can perform quite competitively. One important reason is that the latest off-the-shelf 2D pose estimators, \eg,~CPM \cite{wei2016cpm}, Stacked Hourglass Network \cite{newell2016stacked} can provide 2D pose estimation with high accuracy. On top of this simple 2D pose to 3D pose estimation network \cite{martinez2017a}, Fang \etal \cite{fang2018learning} propose a pose grammar to learn to refine the 3D pose using bi-directional RNNs, which is designed to explicitly incorporate a set of knowledge regarding human body configuration. To exploit the temporal information for 3D human pose estimation, Hossain \etal \cite{hossain2018exploiting} incorporate a sequence-to-sequence regression model using recurrent neural network. Pavllo \etal \cite{pavllo20183d} use dilated temporal convolutions to capture long-term information. 

However, for 3D pose estimation, it is still challenging to handle poses of diverse viewpoints and the body parts of high flexibility. In the task of skeleton based human action recognition \cite{zhang2017view}, some attempts for reducing the effect of viewpoint diversity have been done. However, addressing the viewpoint variation challenge for 3D pose estimation is still overlooked and remains an open problem. In this work, we design a view invariant module to enhance the 3D pose estimation performance. 

\subsection{Adversarial Learning}

Goodfellow \etal \cite{goodfellow2014generative} propose Generative Adversarial Networks (GAN) to help learn efficient generative models via an adversarial process. They simultaneously train two models: a generative model $G$ that captures the data distribution, and a discriminative model $D$ that estimates the probability that a sample comes from the training data of $G$. Motivated by this seminal work, adversarial approaches have been widely applied in various fields \cite{mirza2014conditional,reed2016generative}.

Some recent works \cite{chou2017self,chen2017adversarial,kanazawa2018end} adopt adversarial learning to encourage the deep network to acquire plausible human body configurations. For 2D pose estimation, Chou \etal \cite{chou2017self} and Chen \etal \cite{chen2017adversarial} design discriminators to distinguish groudtruth from generated ones by keypoint heatmaps, and 2D pose, respectively. Kanazawa \etal \cite{kanazawa2018end} add a discriminator to their 3D human body recovery network to determine whether the parameters of the generated 3D mesh correspond to real human bodies or not. 

For 3D pose estimation, adding supervision on each joint (e.g., $l_2$ loss) is widely used for optimizing the estimator without considering the high level human body configurations. This inevitably generates some implausible poses. Yang \etal \cite{yang20183d} design a multi-source discriminator to distinguish the predicted 3D poses from the ground-truth. Three information sources, image, geometric descriptor, as well as the heatmaps and depth maps are utilized for discrimination. In this work, we use a simple view invariant discriminator with the 3D poses of consistent views as input to distinguish the fake 3D poses from real ones. Making distinction under consistent views can reduce the difficultly of adversarial learning. 

%% file: method.tex
\section{Proposed View Invariant Model}

As shown in Figure \ref{fig:pipeline}, our framework consists of three modules: a base 3D pose estimation network, a view-invariant hierarchy refinement/correction network (VI-HC), and a view-invariant discriminator (VID). A base 3D pose estimation module is used to generate initial estimated 3D pose from 2D joints locations. The proposed VI-HC refines the estimated 3D poses under consistent views with global and local transformations. A view-invariant discriminator is used to enhance the performance of the generator, \textit{i.e.} our pose estimator, by enforcing high-level constraints under consistent views during training.  

Generally, the mapping function from a 2D pose $\mathbf{\hat{P}} \in \mathbb{R}^{2 \times J}$ to 3D pose $\mathbf{\hat{S}} \in \mathbb{R}^{3 \times J}$ can be formulated as:
\begin{equation}
\label{equ:mapping}
\mathbf{\hat{S}} = f(\mathbf{\hat{P}}, \theta),
\end{equation}
where $ \theta $ denotes the learnable parameters of the model function $ f $, and $ J $ denotes the number of joints. The objective of the whole model is to estimate each 3D pose $ \mathbf{\hat{S}} $ as close to the ground-truth 3D pose $ \mathbf{S} $ as possible.

\subsection{Base 3D Pose Estimation Network}

Networks that can provide 3D pose estimation could be taken as our base networks. To demonstrate the robustness of our proposed view invariant modules, we take two powerful, representative but simple networks as proposed by \cite{martinez2017a} and \cite{hossain2018exploiting} as our base networks, respectively. For the two base networks, one is designed for individual frame 3D pose estimation while the other is designed for a sequence of 3D pose estimation by exploiting temporal information.
  
\textbf{Baseline-1}. This is a simple network proposed by Martinez \etal \cite{martinez2017a} and we use it to obtain our initial 3D pose from the 2D joint locations input. Note that the 2D joint locations input is obtained from 2D pose estimator, {\it{i.e.}}, Hourglass. This base network is built by stacking two fully connection blocks with residual connection. Each block consists of several linear fully connected layers, followed by batch normalization, a dropout layer and a ReLU activation layer. 
The network encodes the input 2D pose to high dimensional discriminative features first and then projects these representation to 3D space to provide an initial 3D pose $\mathbf{\hat{S}}_{init}$. We represent the coordinate of the $j$-th joint of $ \mathbf{\hat{S}}_{init}$ as  $\mathbf{\hat{s}}_{init}^j \in \mathbb{R}^{3}$. 

\textbf{Baseline-2}. This is a simple network proposed by Hossain \etal \cite{hossain2018exploiting} and we use it to obtain our initial 3D poses from a sequence of 2D joint locations inputs. To exploiting the temporal information across a sequence of 2D joint locations to estimate 3D poses, a sequence-to-sequence LSTM network with residual connections on the decoder side is designed. 

\subsection{View Invariant Hierarchical Correction} 
\begin{figure}[ptb]
	\includegraphics[width=0.47\textwidth]{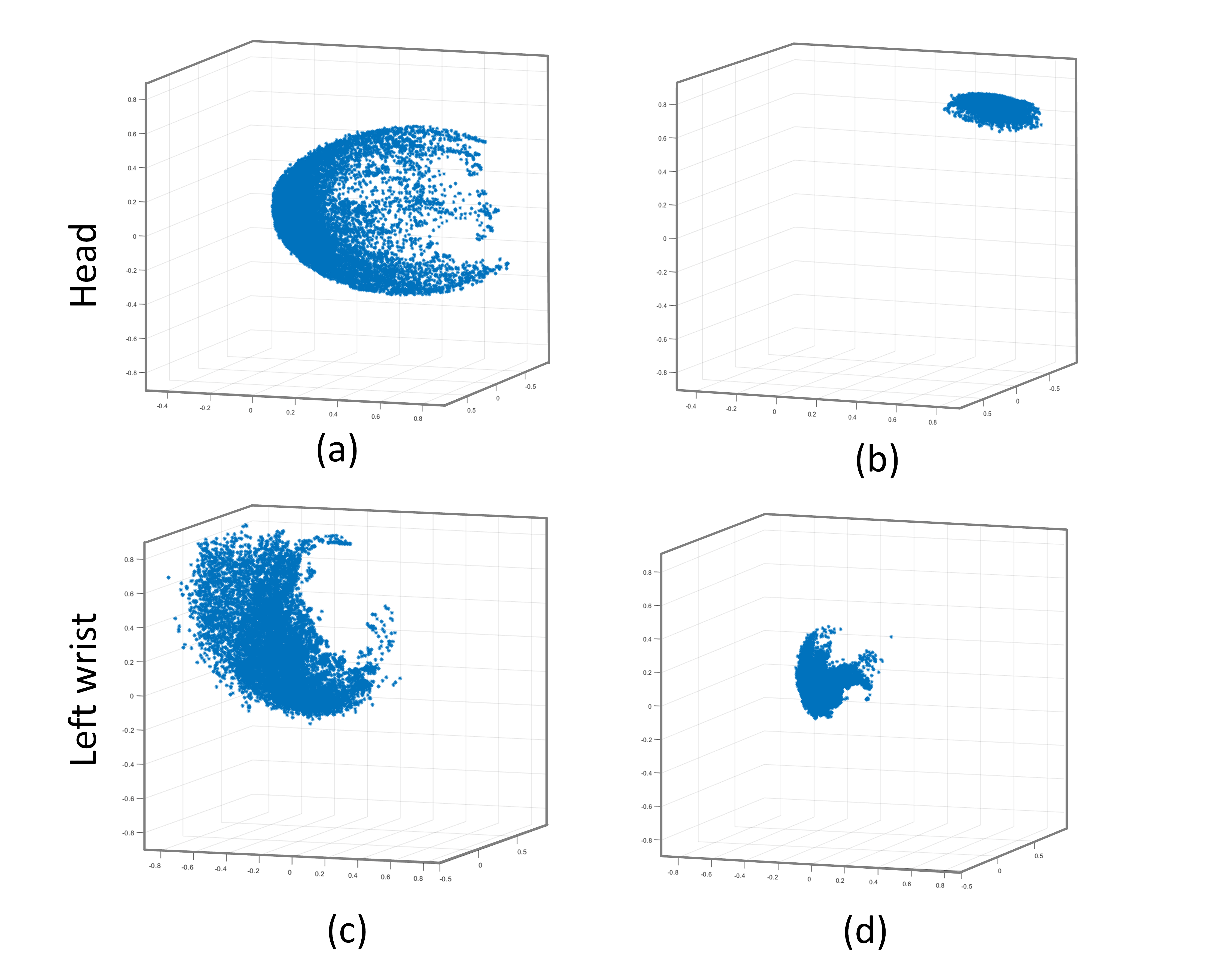}
	\caption{
		Location distributions of different joints before and after transformation. The distribution of head joint (with the root joint located in the coordinate original) (a) before and (b) after global transformation. (c) After global transformation, the distribution of the flexible joint of left wrist is still scattered (with the left shoulder joint located in the coordinate original). (d) After the local transformation for the left arm part, the distribution of left wrist is more concentrated (with the left shoulder joint located in the coordinate original). 
	}
	\label{fig:stat}
\end{figure}
In real life, 3D human poses present great diversity in viewpoints. Such diversity results in scattered distribution for a joint as the examples in Figure \ref{fig:stat} and requires a more powerful 3D pose estimation model to handle all the poses with diverse views. To that end, we propose a view-invariant hierarchical correction module which consists of a global body correction stage and a local body parts correction stage as illustrated in Figure \ref{fig:pipeline} (b) and (c). 

In the global body correction stage, a global transformation module transforms the initial 3D poses that come from the base network to a consistent view, followed by a refinement subnetwork. Similarly, in the local body parts correction stage, we divide the body into five parts with a partition manner similar to that of previous works \cite{chu2016structure,ghezelghieh2016learning,fang2018learning}. For each part, a local part transformation module transforms that body part to a consistent view for further refinement.

\begin{figure}[t]
	\begin{center}
	\includegraphics[width=0.4\textwidth]{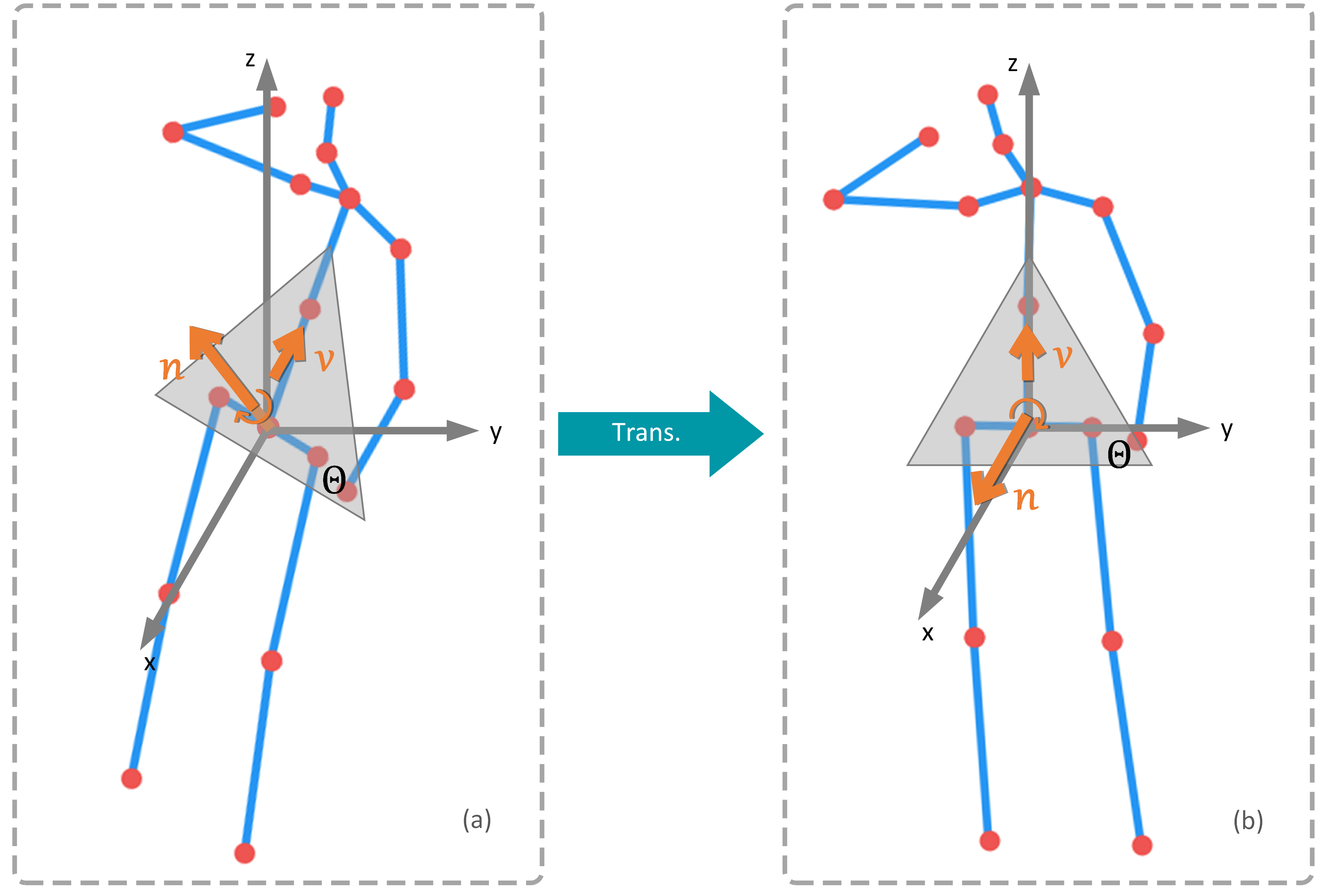}
    \end{center}
	\caption{
		Illustration of transformation for global body correction. 
	}
	\label{fig:explain_rotation}
\end{figure}

\noindent
\textbf{Global body correction.} In the global body correction stage, we focus on correcting 3D poses at consistent viewpoints with respect to the global bodies. The body transformation module transforms body pose to a consistent view, \textit{e.g.}, facing towards the camera with the upper body being upright, as illustrated in Figure \ref{fig:explain_rotation}. 

Specially, we define a plane $ \Theta $ which is determined by the locations of three joints - left/right hips and chest. The root point which is the middle of the left and right hips is defined as the coordinate origin. Within the network, a transformation is performed to make the norm vector $ \mathbf{n} $ of $ \Theta $ parallel with the $ X $ axis and the vector from root point to chest $\mathbf{v}$ parallel with the $ Z $ axis, Y axis is then parallel with $ (\textbf{v} \times \textbf{n})$. $ \mathbf{n} $ and vector $ \mathbf{v} $ are used to determine the rotation angles $ \psi_{X}, \psi_{Y}, \psi_Z $ with respect to $ X, Y, Z $ axes. The transformed poses are facing the camera with the upper body being upright.

The transformation of a global body in the 3D space is formulated as:
\begin{eqnarray}
\mathbf{\tilde{s}}_{gc}^j = \mathbf{R} \times \mathbf{\hat{s}}_{init}^j,  j=1,2,\dots,J,
\end{eqnarray}
where $\mathbf{\tilde{s}}_{gc}^j$ denotes the coordinate of the $j$-th joint after the transformation, and $ \mathbf{R} = \mathbf{R}^X \times \mathbf{R}^Y \times \mathbf{R}^Z $. $ \mathbf{R}^X$, $ \mathbf{R}^Y$, and $ \mathbf{R}^Z$ denote the rotation matrices along the $ X$, $Y$, and $Z$ axes, with rotation angles of $ \psi_X $, $\psi_Y$, $\psi_Z$, respectively. For example, $ \mathbf{R}^X$ can be represented as:
\begin{eqnarray}
\mathbf{R}^X = 
\begin{bmatrix}
1 & 0 & 0 \\
0 & \cos{\psi_X} & \sin{\psi_X} \\
0 & -\sin{\psi_X} & \cos{\psi_X} \\
\end{bmatrix}.
\end{eqnarray}
Note that $\psi_X$,$\psi_Y$,$\psi_Z$ are obtained from vectors $ \mathbf{n} $ and $ \mathbf{v}$.

After the transformation on the initial 3D poses, global body refinement using a fully connected linear subnetwork is performed under such consistent view to effectively correct the 3D poses. \\

\noindent
\textbf{Local body parts correction.} Human bodies are non-rigid and very flexible. Global body transformation can reduce the viewpoint diversity and make the distribution of a joint more concentrated (see Figure \ref{fig:stat}~(b) verse (a)). However, there is still significant flexibility and viewpoint diversity for some joints on local body parts. As illustrated in Figure \ref{fig:stat}~(c), some flexible joints like wrist still have scattered distribution after global body transformation. 

To further reduce the viewpoint diversity with respect to each body part and achieve view invariant, we propose local body part transformations for five body parts, {\it{i.e.}}, left/right arms, left/right legs, and chest-thorax-jaw-head joint part, respectively. After the local body part transformations, the views are much more consistent and thus the location distribution of each joint is more concentrated, as shown in Figure \ref{fig:stat} (d) versus (c). For the $k^{th}$ body part, similar to the global transformation, we take three joints to from a plane $\Theta_k$ and get the norm vector $\mathbf{n}_k$. Then, a transformation is performed to make the norm vector $\mathbf{n}_k$ parallel to the $X$ axis, and the vector $\mathbf{v}_k$ formed by two joints (\eg, shoulder and elbow joints for the arm part) parallel to the $Z$ axis. Similarly, the transformation parameters are obtained from vectors $ \mathbf{n}_k $ and $ \mathbf{v}_k$. Note that for the arm part, upper arm (shoulder and elbow joints) is taken as the sub-part to form the vector $\mathbf{v}_k$. For the leg part, upper leg (hip and knee joints) is taken as the sub-part to form the vector $\mathbf{v}_k$. For the chest-thorax-jaw-head joint chain part, the connection of the chest and thorax joints is taken as the sub-part to form the vector $\mathbf{v}_k$.

For each body part, a refinement subnetwork similar to that for global body refinement is used to further correct that part under a consistent view.

Note that after the local body parts correction, the refined five parts are inversely transformed back based on the transform parameters and combined to have a full body pose. Similarly, based on the global transform parameters, the full body is finally inversed transformed back to the original view to obtain the final 3D pose as illustrated in Figure \ref{fig:pipeline}.

\subsection{View Invariant Discriminator Module}
We encourage our pose estimator to explore prior human body configurations to avoid implausible 3D poses by adversarial learning. Considering it is easier to learn patterns from data of consistent viewpoints than that of diverse viewpoints, we design a simple discriminator to distinguish generated 3D poses from real 3D poses under consistent views.  

During adversarial learning, our 3D pose estimator which is composed of the base network and view-invariant hierarchical correction network is taken as the generator $G$. The estimated 3D poses are treated as ``fake" samples (label 0) while the groudtruth 3D poses as ``real" samples (label 1) to train the discriminator. The goal of $G$ is to generate poses with the distribution being similar to that of the groundtruth 3D poses, intending that the discriminator cannot differentiate between real samples and the estimated poses. 

We propose to conduct the adversarial learning using the 3D poses under consistent viewpoints rather than the original views as illustrated in Figure 1. The more concentrated of the pose distribution by removing viewpoint variation, the easier to optimize the discriminator. Otherwise, the discriminator needs to be able to distinguish fake from real samples for various views, which is more challenging.

As the generator $G$, the pose estimator tries to predict poses as real as possible to fool discriminator $D$ by optimizing the following additional loss:
\begin{eqnarray}
\label{l_g}
\mathcal{L}_G = \sum \mathcal{L}_{bce}(D(G(\hat{\mathbf{P}})), 1),
\end{eqnarray}
where $\mathcal{L}_{bce}$ is the binary cross entropy loss. $G(\hat{\mathbf{P}})$ denotes the estimated 3D pose under the consistent view (ahead the \emph{Inv. Body Trans.} as shown in Figure \ref{fig:pipeline}). $D(\cdot)$ represents the classification score of the discriminator.

The loss for training discriminator $D$ is 
\begin{eqnarray}
\mathcal{L}_D = \sum \mathcal{L}_{bce}((D(\mathbf{S}), 1) + \mathcal{L}_{bce}(D(G(\mathbf{\hat{P}})), 0),
\end{eqnarray}
where $\mathbf{\tilde{S}}$ denotes the groudtruth 3D pose under consistent view, {\it{i.e.}}, after global body transformation.


\subsection{Joint Learning}

Our designed view invariant 3D pose estimation network is an end-to-end network. We jointly train the network based on the loss
\begin{eqnarray}
\mathcal{L} = \mathcal{L}_{pose} + \lambda\mathcal{L}_G,
\end{eqnarray}
where $\mathcal{L}_{pose}$ is the $l_2$ loss on the joints for both the global body correction and local body parts correction stages, while $\mathcal{L}_G$ is the cross-entropy loss from the discriminator, and $\lambda$ is a hyperparameter which is experimentally determined. 

%% file: experiments.tex
\section{Experiments}

\subsection{Datasets}

Similar to previous works \cite{zhou2017towards,sun2017compositional,martinez2017a}, we conduct our experiments on the popular Human3.6M dataset~\cite{ionescu2014human}. Human3.6M is currently the largest publicly available dataset for 3D human pose estimation. This dataset consists of 3.6 million images where 7 professional actors perform 15 everyday activities such as purchasing, walking and sitting down. Four cameras are used. We also show qualitative results on the MPII 2D pose estimation dataset \cite{andriluka2014142d}, for which the ground truth 3D is not available and the images are captured in the wild. 

\subsection{Evaluation Metrics and Protocols}

To fully validate the effectiveness of our proposed method, we use several commonly used evaluation metrics as follows.

\textit{Joint Error:} the \textit{mean per joint position error} (MPJPE). Most previous 3D pose estimation works use this metric \cite{zhou2017towards,sun2017compositional,martinez2017a,tekin2016direct,yasin2016a,nie2017monocular}.

\textit{PA Joint Error:} the MPJPE is calculated after aligning the predicted 3D pose and groundtruth 3D pose via a rigid transformation using Procrustes Analysis \cite{gower1975generalized}.

\textit{Bone Error:} the mean per bone position error. It measures the relative joint location accuracy compared with the groundtruth \cite{sun2017compositional}.

\textit{Bone Std:} the bone length standard deviation. It measures the stability of bone length \cite{sun2017compositional}.

We report performance of our proposed method using three protocols.

\textit{Protocol \#1:}  For Human3.6M, the standard Protocol \#1 is to use all four camera views of subjects S1, S5, S6, S7 and S8 for training and the subjects S9 and S11 for testing. We report the MPJPE in millimeters between the groundtruth and our prediction across all joints. 

\textit{Protocol \#2:} Under the same setting as Protocol \#1, the evaluation by PA Joint Error is referred to as Protocol \#2 \cite{martinez2017a,fang2018learning,pavlakos2017volumetric}.

\textit{Protocol \#3:} Due to limited camera views, it is easy for a pose estimation model to overfit to the limited camera views. To validate the generalization ability of a model, Fang {\etal} \cite{fang2018learning} propose a cross-view protocol, where only 3 camera views are used for training while the other one is for testing. We refer to this protocol as Protocol \#3. 

\begin{figure*}[t]
	\includegraphics[width=1.0\textwidth]{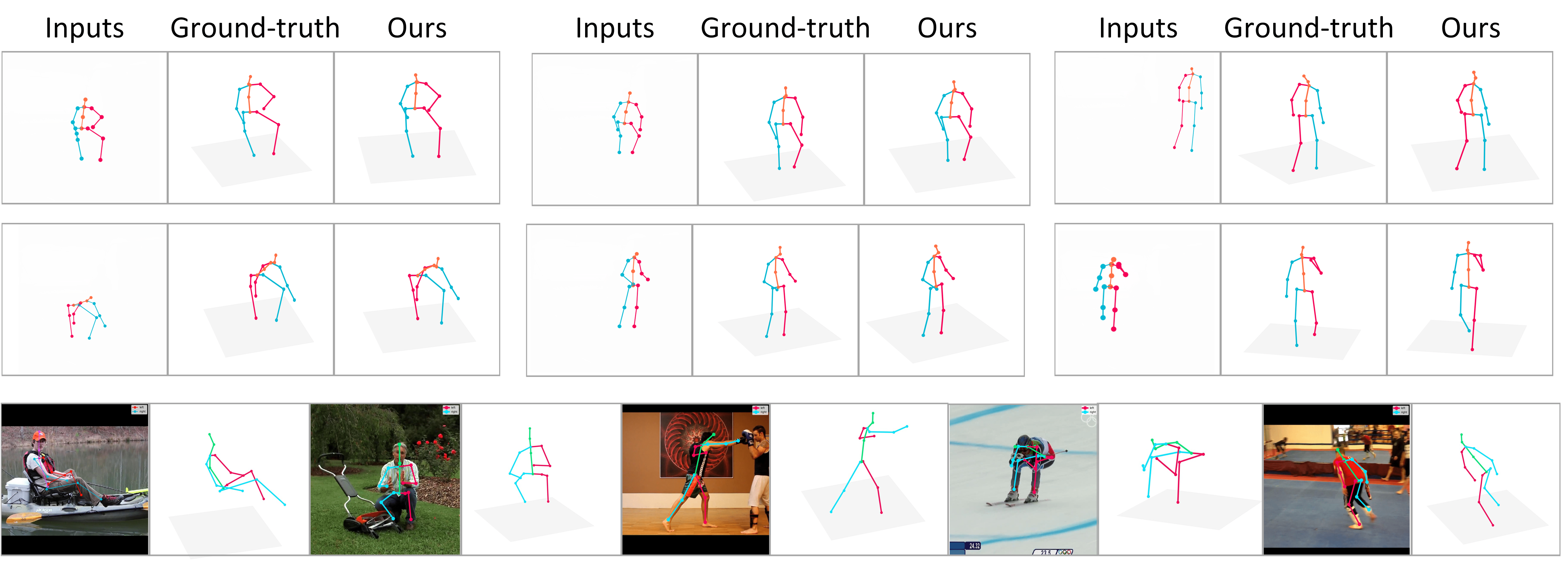}
	\vspace{-6mm}
	\caption{Qualitative results of our proposed method on Human3.6M (first two rows) and MPII (last row). The last example in the second row is a failure case because of the lack of appearance information.}
	\label{fig:qualitative}
\end{figure*}

\subsection{Implementation Details} 

We implement our method using PyTorch \cite{paszke2017automatic}. In our base networks, similar to \cite{martinez2017a,fang2018learning,hossain2018exploiting}, the state-of-the-art stacked hourglass network \cite{newell2016stacked}, pre-trained on MPII and fine-tuned on Human3.6M, is used to estimate 2D poses on Human3.6M. 

Our refinement network \emph{VI-HC} has a similar architecture as the base network \cite{martinez2017a}, where linear block and residual connection are used. Differently, the number of neurons for our fully connected layers are 400 and 800, respectively.
For the discriminator, we adopt a simple encoder design which is composed of four fully connected layers. 
We set $\lambda$ as 0.001. The entire network is end-to-end trained with the base network pretrained first. 


\subsection{Ablation Studies}

\begin{table}[t]
	\small
	\centering
	\begin{tabular}{@{}l|c|c@{}}
		\hline
		Method 			              & Avg. & $ \Delta $	\\
		\hline
		Baseline-1 (Martinez {\etal} ICCV'17) & 62.8 & - \\ 
		B+HC          & 61.8 & -1.0 \\				
		B+VI-LC	                  & 60.2 & -2.6 \\
		B+VI-GC                   & 59.8 & -3.0 \\
		B+VI-HC                   & 58.4 & -4.4 \\
		B+VI-HC-D                    & 57.8 & -5.0 \\
		\textbf{B+VI-HC-VID} (ours)                   & 57.1 & -5.7 \\
		\hline
	\end{tabular}
	\vspace{0.2em}
	\caption{
		Ablation studies of different components in our design on top of Baseline-1 on Human3.6M under Protocol \#1, in terms of MPJPE (mm). 
	}
	\label{tab:ablation_hm36}
\end{table}

\begin{table}[t]
	\small
	\centering
	\begin{tabular}{@{}l|c|c@{}}
		\hline
		Method 			              & Avg. & $ \Delta $	\\
		\hline
		Baseline-1$^{\diamond}$ (Martinez {\etal} \cite{martinez2017a} ICCV'17)  & 74.2 & - \\ 			
		B$^{\diamond}$+VI-HC-VID (ours)   & 68.9 & -5.3 \\ 		
		\hline
	\end{tabular}
	\vspace{0.2em}
	\caption{
		Ablation studies of different components in our design on top of a weaker base model \textit{Baseline-1$^{\diamond}$} on Human3.6M under Protocol \#1, in terms of MPJPE (mm). 
	}
	\label{tab:ablation_hm36-weaker}
\end{table}

To demonstrate the effectiveness of each component of our proposed model, we perform various experiments on Human3.6M under Protocol \#1 and report the MPJPE results in Table \ref{tab:ablation_hm36}. Without loss of generality, we perform the ablation study on top of \emph{Baseline-1} \cite{martinez2017a}.

\begin{itemize}
	\setlength{\itemsep}{0pt}
	\item \textit{Baseline-1}:~the baseline 3D pose estimation network \cite{martinez2017a} which we take as our base network.
	\item \textit{B+HC}:~the \textbf{B}ase network \cite{martinez2017a} followed by \textbf{H}ierarchical \textbf{C}orrection \emph{without} view transformations.		
	\item \textit{B+VI-GC (or B+VI-LC)}:~the \textbf{B}ase network followed by \textbf{V}iew \textbf{I}nvariant \textbf{G}lobal (or \textbf{L}ocal) \textbf{C}orrection only.
	\item \textit{B+VI-HC}:~the \textbf{B}ase network followed by \textbf{V}iew \textbf{I}nvariant \textbf{H}ierarchical \textbf{C}orrection.
	\item \textit{B+VI-HC-D}:~\textit{B+VI-HC} scheme with a \textbf{D}iscriminator during training, where the discriminator operates under the original views.
	\item \textit{B+VI-HC-VID}:~\textit{B+VI-HC} scheme with \textbf{V}iew \textbf{I}nvariant \textbf{D}iscriminator. \textbf{This is our final scheme.} It is the same as \textit{B+VI-HC-D} except that the discriminator operates under the transformed views rather than the original views.
	\item \textit{{Baseline-1}$^{\diamond}$}:~a weaker baseline 3D pose estimation network which uses only one fully connection block rather than two \cite{martinez2017a}.
	\item \textit{B$^{\diamond}$+VI-HC-VID}:~a weaker base network \textit{Baseline-1$^{\diamond}$} followed by our proposed VI-HC and VID.		
	
\end{itemize}

\begin{table*}[t]
	\centering
	\small
	\begin{tabular}{@{}l|c|c|c|c|c|c|c|c@{}}
		\hline
		Metric              & \multicolumn{2}{c}{Joint Error} & \multicolumn{2}{|c}{PA Joint Error} & \multicolumn{2}{|c}{Bone Error}	& \multicolumn{2}{|c}{Bone Std} \\
		\hline
		Method              & Baseline & Ours & Baseline & Ours  & Baseline & Ours & Baseline & Ours\\
		\hline
		Knee(\ra~Hip)       & 64.6 & $ 53.9_{\downarrow10.7} $  & 50.8 & $ 45.5_{\downarrow5.3} $ & 63.4 & $ 57.7_{\downarrow5.7} $ & 21.1 & $ 14.1_{\downarrow7.0} $ \\
		Ankle(\ra~Knee)     & 87.6 & $ 83.4_{\downarrow4.2} $ & 64.7 & $ 60.8_{\downarrow3.9} $ & 83.5 & $ 76.8_{\downarrow6.7} $ & 26.5 & $ 13.8_{\downarrow12.7} $\\
		Wrist(\ra~Elbow)    & 112.9 & $ 99.3_{\downarrow13.6} $ & 81.8 & $ 74.7_{\downarrow7.1} $ & 76.7 & $ 71.9_{\downarrow4.8} $ & 28.9 & $ 21.6_{\downarrow7.3} $ \\
		Elbow(\ra~Shoulder) & 86.4 & $ 75.0_{\downarrow11.4} $ & 57.1 & $ 48.8_{\downarrow8.3} $ & 61.7 & $ 59.3_{\downarrow2.4} $ & 19.8 & $ 16.4_{\downarrow3.4} $ \\
		Shoulder(\ra~Thorax)& 60.7 & $ 54.9_{\downarrow5.8} $ & 35.8 & $ 30.9_{\downarrow4.9} $ & 40.5 & $ 38.1_{\downarrow2.4} $ & 9.2 & $ 7.6_{\downarrow1.6} $ \\
		\hline
		Avg.                & 68.1 & $ 62.2_{\downarrow5.9} $ & 51.3 & $ 44.2_{\downarrow7.1} $ & 51.6 & $ 47.5_{\downarrow4.1} $ & 16.0 & $ 11.9_{\downarrow4.1} $ \\
		\hline
	\end{tabular}
	\vspace{0.2em}
	\caption{
		Detailed results for some joints and bones for \emph{Baseline-1} \cite{martinez2017a} and \emph{Ours} based on Baseline-1, under Protocol \#2 on Human3.6M.}
	\label{table:joint_bone}
\end{table*}

\begin{table}[t] 
	\centering
	\small
	\begin{tabular}{@{}p{2.0cm}|p{0.8cm}p{0.8cm}p{0.8cm}p{0.8cm}|p{0.5cm}}
		\hline
		Bone pairs  & U.Arm & L.Arm & U.Leg & L.Leg & Avg.\\
		\hline
		Baseline & 17.2 & 26.2 & 13.1 & 16.1 & 18.2 \\
		Ours & 11.3 & 17.2 & 6.8 & 6.9 & 10.6\\
		\hline
	\end{tabular}
	\vspace{0.2em}
	\caption{
		Evaluations of the symmetry of limbs for \emph{Baseline-1} \cite{martinez2017a} and \emph{Ours} based on Baseline-1, under Protocol \#2 on Human3.6M.
	}
	\label{tab:symmetry}
\end{table}

As shown in Table \ref{tab:ablation_hm36}, view invariant global body correction (\textit{B+VI-GC}) and view invariant local body parts correction (\textit{B+VI-LC}) reduce the MPJPE by 3.0mm and 2.6mm, respectively. Combining the two level corrections achieve 4.4mm error reduction. Moreover, the adversarial learning under consistent view further decreases the error by 1.3mm, and our final scheme achieves 5.7mm reduction.

\noindent
\textbf{Consistent viewpoint helps effectively refine 3D poses.}~To verify  that correcting the initial 3D poses under consistent viewpoints is more effective, we also evaluate the scheme with two stage 3D pose corrections without view transformations, {\it{i.e.,}} \textit{B+HC}. Compared with \textit{Baseline}, this scheme only reduces the MPJPE by 1.0mm while the correction with view transformations reduces the MPJPE by 4.4mm. There are two main reasons. i) Additional correction introduces more parameters so that it becomes harder to train the model. Martinez \etal \cite{martinez2017a} also report that deeper network dose not improve the performance. ii) The poses  are from various viewpoints. It is not easy for a model to handle them with such high diversity. In contrast, our model with view transformations \textit{B+VI-HC} transforms the poses to consistent views and makes the learning easier. 

\textit{B+VI-GC} produces superior results to \textit{B+VI-LC}. In \textit{B+VI-LC}, each type of body part has a subnetwork which only makes use of the intra part rather than cross part information for refinement. In contrast, \textit{B+VI-GC} takes all the joints as input for refinement. First, more context information (all the joints) is used for the refinement in global correction \textit{B+VI-GC} than that in local part correction \textit{B+VI-LC}. Second, local corrections can only improve the local joint details relative to the part, but have difficulty correcting the position errors of the part as a whole, due to lack of context information. The global errors could have larger contribution to the total error. 

\noindent
\textbf{Consistent viewpoint helps the discriminator.}~If we feed the poses under the original views to the discriminator (\textit{B+VI-HC-D}), the adversarial learning gains 0.6mm compared with the one not using adversarial learning (\textit{B+VI-HC}). In contrast, the gain increases to 1.3mm if we feed the poses transformed to consistent views to the discriminator (\textit{B+VI-HC-VID}). 
The poses with consistent views can help better train the discriminator and further improve the performance of the pose estimator.

To validate the robustness of our view invariant design on different base networks, we also conduct experiments on a weaker base network, which is built using only one fully connection block rather than two \cite{martinez2017a}, referred to as, \textit{Baseline-1$^{\diamond}$}. Similarly, our view invariant scheme achieves an error reduction of 5.3mm as shown in Table \ref{tab:ablation_hm36-weaker}.

\begin{table*}[t]
	\centering
	\setlength{\tabcolsep}{4pt}
	\resizebox{\textwidth}{!}{
		\begin{tabular}{@{}lccccccccccccccc|c@{}}
			\midrule
			\textbf{Method} & Direct. & Discuss & Eating & Greet & Phone & Photo & Pose & Purch. & Sitting & SittingD. & Smoke & Wait & WalkD. & Walk & WalkT. & Avg.\\
			\midrule
			Zhou~{\etal}\cite{zhou2016sparseness} (CVPR'16) & 87.4 & 109.3 & 87.1 & 103.2 & 116.2 & 143.3 & 106.9 & 99.8 & 124.5 & 199.2 & 107.4 & 118.1 & 114.2 & 79.4 & 97.7 & 113.0\\
			Du~{\etal}\cite{du2016marker} (ECCV'16) & 85.1 & 112.7 & 104.9 & 122.1 & 139.1 & 135.9 & 105.9 & 166.2 & 117.5 & 226.9 & 120.0 & 117.7 & 137.4 & 99.3 & 106.5 & 126.5\\
			Park~{\etal}\cite{park20163d} (ECCVW'16) & 100.3 & 116.2 & 90.0 & 116.5 & 115.3 & 149.5 & 117.6 & 106.9 &  137.2 & 190.8 & 105.8 & 125.1 & 131.9 & 62.6 & 96.2 & 117.3\\
			Pavlakos~{\etal}\cite{pavlakos2017volumetric} (CVPR'17) & 67.4 & 71.9 & 66.7 & 69.1 & 72.0 & 77.0 & 65.0 & 68.3 & 83.7 & 96.5 & 71.7 & 65.8 & 74.9 & 59.1 & 63.2 & 71.9\\
			Zhou~{\etal}\cite{zhou2017towards} (ICCV'17) & 54.8 & 60.7 & 58.2 & 71.4 & 62.0 & 65.5 & 53.8 & 55.6 & 75.2 & 111.6 & 64.1 & 66.0 & 51.4 & 63.2 & 55.3 & 64.9\\
			Sun~{\etal}\cite{sun2017compositional} (ICCV'17) & 52.8 & 54.8 & 54.2& 54.3 & 61.8 & 53.1 & 53.6 & 71.7 & 86.7 & 61.5 & 67.2 & 53.4 & 47.1 & 61.6 & 53.4 & 59.1\\
			Fang~{\etal}\cite{fang2018learning} (AAAI'18) & 50.1 & 54.3 & 57.0 & 57.1 & 66.6 & 73.3 & 53.4 & 55.7 & 72.8 & 88.6 & 60.3 & 57.7 & 62.7 & 47.5 & 50.6 & 60.4 \\
			Hossain~{\etal}\cite{hossain2018exploiting} (ECCV'18) & 48.4 & 50.7 & 57.2 & 55.2 & 63.1 & 72.6 & 53.0 & 51.7 & 66.1 & 80.9 & 59.0 & 57.3 & 62.4 & 46.6 & 49.6 & 58.3\\
			Pavlakos~{\etal}\cite{pavlakos2018ordinal} (CVPR'18) (wo/ Ord)  & -- & -- & -- & -- & -- & -- & -- & -- & -- & -- & -- & -- & -- & -- & -- & 59.1\\			
			Pavlakos~{\etal}\cite{pavlakos2018ordinal}* (CVPR'18)  & 48.5 & 54.4 & 54.4 & 52.0 & 59.4 & 65.3 & 49.9 & 52.9 & 65.8 & 71.1 & 56.6 & 52.9 & 60.9 & 44.7 & 47.8 & 56.2 \\
			Yang~{\etal}\cite{yang20183d} (CVPR'18) & 51.5 & 58.9 & 50.4 & 57.0 & 62.1 & 65.4 & 49.8 & 52.7 & 69.2 & 85.2 & 57.4 & 58.4 & 43.6 & 60.1 & 47.7 & 58.6 \\
			
			\midrule
			Martinez~{\etal}\cite{martinez2017a} (ICCV'17) (Baseline-1) & 51.8 & 56.2 & 58.1 & 59.0 & 69.5 & 78.4 & 55.2 & 58.1 & 74.0 & 94.6 & 62.3 & 59.1 & 65.1 & 49.5 & 52.4 & 62.9\\			
			Ours~(with Baseline-1) & \underline{46.6} & \underline{54.0} & \underline{55.1}  & \underline{55.2} & \underline{61.4} & \underline{69.8} & \underline{52.0} & \underline{52.6} & \underline{68.1} & \underline{75.0} & \underline{56.7} & \underline{56.0} & \underline{60.5} & \underline{44.5} & \underline{48.7} & \underline{57.1} \\
			\midrule
			Hossain~{\etal}\cite{hossain2018exploiting} (ECCV'18) (Baseline-2) & \underline{48.4} & 50.7 & 57.2 & 55.2 & 63.1 & 72.6 & 53.0 & 51.7 & 66.1 & 80.9 & 59.0 & 57.3 & 62.4 & 46.6 & 49.6 & 58.3 \\
			Ours~(with Baseline-2) & 48.5 & \underline{49.5}  & \underline{55.0} & \underline{52.5} & \underline{62.1} & \underline{69.5} & \underline{52.7} & \underline{49.6} & \underline{63.9} & \underline{76.6} & \underline{57.4} & \underline{55.8} & \underline{60.3} & \underline{46.5} & \underline{49.3}
			& \underline{56.6} \\
			\midrule
		\end{tabular}
	}
	\caption{
		Comparisons on Human3.6M under \textbf{Protocol \#1} in terms of MPJPE (mm). The underlined numbers represent the better results between ours and the baseline. Note that the 2D inputs are obtained with a fine-tuned stacked hourglass 2D pose detector for both Baselines and our schemes. For the work Pavlakos~{\etal}\cite{pavlakos2018ordinal} marked by (*), additional annotations of the ordinal depth on the 2D human pose datasets are utilized. Pavlakos~{\etal}\cite{pavlakos2018ordinal} denotes the results without using ordinal depth annotations.
	}		
	\label{tab:h36mprotocol1}
\end{table*}

\begin{table*}[t]
	\centering
	\setlength{\tabcolsep}{4pt}
	\resizebox{\textwidth}{!}{
		\begin{tabular}{@{}lccccccccccccccc|c@{}}
			\midrule
			\textbf{Method} & Direct. & Discuss & Eating & Greet & Phone & Photo & Pose & Purch. & Sitting & SittingD. & Smoke & Wait & WalkD. & Walk & WalkT. & Avg.\\
			\midrule
			Zhou~{\etal}\cite{zhou2016sparseness} (CVPR'16) & 99.7 & 95.8 & 87.9 & 116.8 & 108.3 & 107.3 & 93.5 & 95.3 & 109.1 & 137.5 & 106.0 & 102.2 & 106.5 & 110.4 & 115.2 & 106.7\\
			Bogo~{\etal}\cite{bogo2016keep} (ECCV'16) & 62.0 & 60.2 & 67.8 & 76.5 & 92.1 & 77.0 & 73.0 & 75.3 & 100.3 & 137.3 & 83.4 & 77.3 & 86.8 & 79.7 & 87.7 & 82.3\\
			Nie~{\etal}\cite{nie2017monocular} (ICCV'17) & 62.8 & 69.2 & 79.6 & 78.8 & 80.8 & 72.5 & 73.9 & 96.1 & 106.9 & 88.0 & 86.9 & 70.7 & 71.9 & 76.5 & 73.2 & 79.5\\
			Moreno-Noguer\cite{moreno20173d} (CVPR'17) & 66.1 & 61.7 & 84.5 & 73.7 & 65.2 & 67.2 & 60.9 & 67.3 & 103.5 & 74.6 & 92.6 & 69.6 & 71.5 & 78.0 & 73.2 & 74.0\\
			Pavlakos~{\etal}\cite{pavlakos2017volumetric} (CVPR'17) & -- & -- & -- & -- & -- & -- & -- & -- & -- & -- & -- & -- & -- & -- & -- & 51.9\\
			Fang~{\etal}\cite{fang2018learning} (AAAI'18) & 38.2 & 41.7 &	43.7 & 44.9 & 48.5 & 55.3 & 40.2 & 38.2 & 54.5 & 64.4 &	47.2 & 44.3 & 47.3 & 36.7 &	41.7 & 45.7\\
			Hossain~{\etal}\cite{hossain2018exploiting} (ECCV'18)  & 35.7 & 39.3 & 44.6 & 43.0 & 47.2 & 54.0 & 38.3 & 37.5 & 51.6 & 61.3 & 46.5 & 41.4 & 47.3 & 34.2 & 39.4 & 44.1\\
			Pavlakos~{\etal} \cite{pavlakos2018ordinal}* (CVPR'18) & 34.7 & 39.8 & 41.8 & 38.6 & 42.5 & 47.5 & 38.0 & 36.6 & 50.7 & 56.8 & 42.6 & 39.6 & 43.9 & 32.1 & 36.5 & 41.8 \\
			Yang~{\etal}\cite{yang20183d} (CVPR'18) & 26.9 & 30.9 & 36.3 & 39.9 & 43.9 & 47.4 & 28.8 & 29.4 & 36.9 & 58.4 & 41.5 & 30.5 & 29.5 & 42.5 & 32.2 & 37.7 \\
			\midrule
			Martinez~{\etal}\cite{martinez2017a} (ICCV'17) (Baseline-1) & 39.5 & 43.2&46.4&	47.0&	51.0&	56.0&	41.4&	40.6&	56.5&	69.4&	49.2&	45.0&	49.5&	38.0&	43.1&	47.7\\			
			Ours~(with Baseline-1) & \underline{35.8} & \underline{41.0} & \underline{42.4} & \underline{44.1} & \underline{45.9} & \underline{50.6} & \underline{39.5} & \underline{37.7} & \underline{52.2} & \underline{56.6} & \underline{45.5} & \underline{41.7} & \underline{46.6} & \underline{33.7} & \underline{38.6} & \underline{43.4} \\ 
			\midrule
			Hossain~{\etal}\cite{hossain2018exploiting} (ECCV'18) (Baseline-2) & \underline{35.7} & 39.3 & 44.6 & 43.0 & 47.2 & 54.0 & 38.3 & 37.5 & 51.6 & 61.3 & 46.5 & 41.4 & 47.3 & \underline{34.2} & 39.4 & 44.1 \\
			Ours~(with Baseline-2) & 35.9 & \underline{38.2} & \underline{42.8} & \underline{41.9} & \underline{45.6} & \underline{51.5} & \underline{38.1} & \underline{36.9} & \underline{50.1} & \underline{58.1} & \underline{45.5} & \underline{39.6} & \underline{45.2} & 34.6 & \underline{39.1} & \underline{42.8}\\		
			\midrule
		\end{tabular}
	}
	
	\caption{
		Comparisons on Human3.6M under \textbf{Protocol \#2} in terms of MPJPE (mm) using \textit{PA Joint Error} metric. Note that for the work Pavlakos~{\etal}\cite{pavlakos2018ordinal} marked by (*), additional annotations of the ordinal depth on the 2D human pose datasets are utilized.
	}
	\label{tab:h36mprotocol2}
\end{table*}

\begin{table*}[tp]
	\centering
	\setlength{\tabcolsep}{4pt}
	\resizebox{\textwidth}{!}{
		\begin{tabular}{@{}lccccccccccccccc|c@{}}
			\midrule
			\textbf{Method} & Direct. & Discuss & Eating & Greet & Phone & Photo & Pose & Purch. & Sitting & SittingD. & Smoke & Wait & WalkD. & Walk & WalkT. & Avg.\\
			\midrule
			Pavlakos~{\etal}\cite{pavlakos2017volumetric} (CVPR'17) & 79.2 & 85.2 & 78.3 & 89.9 & 86.3 & 87.9 & 75.8 & 81.8 & 106.4 & 137.6 & 86.2 & 92.3 & 72.9 & 82.3 & 77.5 & 88.6\\
			Bie~{\etal}\cite{nie2017monocular} (ICCV'17) & 103.9 & 103.6 & 101.1 & 111.0 & 118.6 & 105.2 & 105.1 & 133.5 & 150.9 & 113.5 & 117.7 & 108.1 & 100.3 & 103.8 & 104.4 & 112.1\\
			Zhou {\etal}\cite{zhou2017towards} (ICCV'17) & 61.4 & 70.7 & 62.2 & 76.9 & 71.0 & 81.2 & 67.3 & 71.6 & 96.7 & 126.1 & 68.1 & 76.7 & 63.3 & 72.1 & 68.9 & 75.6\\
			Fang {\etal}\cite{fang2018learning} (AAAI'18) & 57.5 &  57.8  & 81.6 & 68.8 &	75.1 & 85.8 & 61.6 & 70.4 &	95.8 & 106.9 & 68.5 & 70.4 & 73.8 & 58.5 & 59.6 & 72.8\\
			\midrule
			Martinez {\etal}\cite{martinez2017a} (ICCV'17) (Baseline-1) & 65.7 & 68.8 & 92.6 & 79.9 & 84.5 & 100.4 & 72.3 & 88.2 & 109.5 & 130.8 & 76.9 & 81.4 & 85.5 & 69.1 & 68.2 & 84.9\\			
			Ours~(with Baseline-1) & \underline{56.4} & \underline{60.9} & \underline{69.1} & \underline{70.0} & \underline{72.4} & \underline{84.1} & \underline{60.3} & \underline{71.3} & \underline{82.9} & \underline{89.0} & \underline{67.1} & \underline{70.9} & \underline{74.0} & \underline{66.4} & \underline{65.2} & \underline{70.8} \\
			\midrule
		\end{tabular}
	}
	
	\caption{
		Comparisons on Human3.6M under \textbf{Protocol \#3} in terms of MPJPE (mm). Samples of three camera views are used for training and those of the other one are used for testing.}
	\label{tab:h36mprotocol3}
\end{table*}

\begin{figure}[t]
	\centering
	\includegraphics[width=0.4\textwidth]{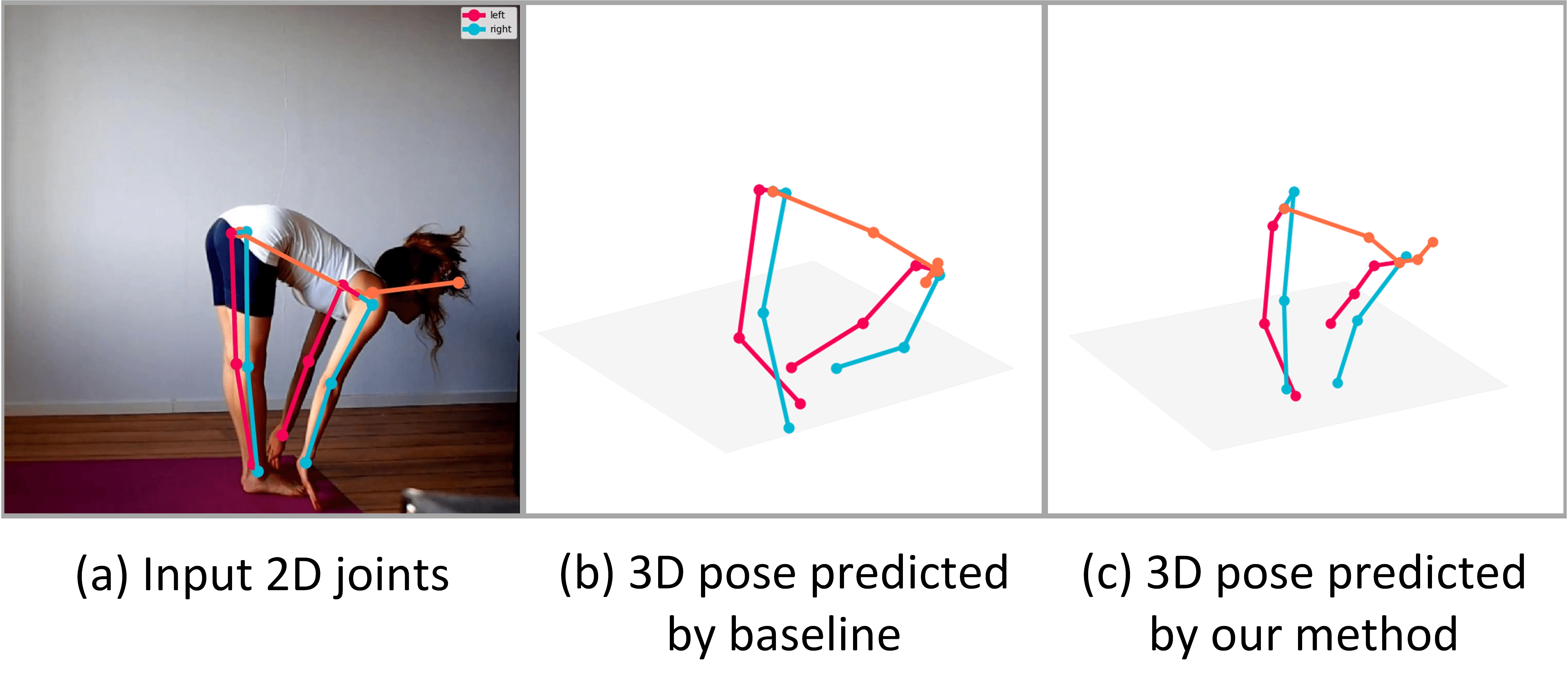}
	\vspace{-3mm}
	\caption{Our method can correct the implausible pose.}
	\label{fig:plausible}
\end{figure}

\subsection{Plausibility Analysis}

Table \ref{table:joint_bone} shows the performance improvement of our proposed method for all joints and bones. The columns with results of \emph{Joint Error} and \emph{PA Joint Error} show that for the joints far away from the root joint (\eg, wrist, elbow), it is more difficult to estimate, since these joints are more flexible and have more scattered spatial distributions. Our proposed method dramatically improves the accuracy of these joints thanks to the hierarchical correction design, in which the view transformations reduce the diversity of viewpoints and easy the refinements. We have the similar observations on the bone errors ({\it{e.g.}}, in terms of \emph{Bone Error} and \emph{Bone Std}), where the bones in four limbs have larger errors.

To evaluate the robustness of our scheme in terms of body structure preservation, in Table \ref{tab:symmetry}, we present the symmetry evaluation results of \emph{Baseline} and \emph{Ours}. The symmetry metric is defined as the difference between the left and right limb lengths. The smaller of the difference, the better of the preservation of the body structure. Our method predicts plausible 3D poses with more symmetrical structures.

We show some visualization results on Human3.6M and MPII in Figure \ref{fig:qualitative}. The results show that our method predicts plausible 3D pose well. We also show our model can correct some implausible poses thanks to the view invariant correction and adversarial learning. Figure \ref{fig:plausible} shows that the baseline model estimates pose with wrong body structure, while our model provides much better prediction.

\subsection{Comparison with the State-of-the-arts}

We compare our schemes based on the two base networks with the state-of-the-art approaches on Human3.6M.

Table \ref{tab:h36mprotocol1} presents the comparisons under Protocol \#1. Our proposed method achieves an MPJPE of 57.1 \textit{mm}. Compared with the powerful baseline networks \cite{martinez2017a} as marked by \emph{Baseline-1} and \emph{Baseline-2}, our schemes achieve about \textbf{9\% } and 3\% improvements and obtains gains even on all the action classes. For some challenging actions with diverse postures and viewpoints like \emph{sitting down}, our method gains about 21\% over \emph{Baseline-1}.
These results validate that our proposed model is very efficient by taking the viewpoint consistency into account. Based on \emph{Baseline-2} \cite{hossain2018exploiting}, our scheme achieves the best performance (MPJPE 56.5 \textit{mm}) among the state-of-the-art approaches except Pavlakos~{\etal}\cite{pavlakos2018ordinal}* (CVPR'18), which uses additional annotations of the ordinal depth on the 2D human pose datasets. Our performance is superior to that of Pavlakos~{\etal}\cite{pavlakos2018ordinal}* (CVPR'18) (wo/ Ord) when it does not use these additional annotations. Note that the relative gain on \emph{Baseline-2} is smaller than that on \emph{Baseline-1} because the higher performance of a baseline, the harder to obtain gain with a smaller improvement space.   


Under Protocol \#2, as shown in Table \ref{tab:h36mprotocol2}, our proposed method performs best for all the action classes, compared with our baseline model. The improvement is about 9\% over \emph{Baseline-1} \cite{martinez2017a} and 3\% over \emph{Baseline-2}. Our performance is inferior to that of Yang \etal \cite{yang20183d}. One possible reason is that our base networks take off-line obtained 2D pose to estimate the 3D pose and lose the opportunity to further exploit the image information. Our performance is comparable to that of the other state-of-the-art approaches.  


Under Protocol \#3, as shown in Table \ref{tab:h36mprotocol3}, our model improves performance significantly, {\it{i.e.}}, by \textbf{16\%}, compared with \emph{Baseline-1} \cite{martinez2017a}, and outperforms the recently published best results. Our scheme is robust to inputs of unseen viewpoints, since the initial poses are transformed to consistent views within the network.


%% file: conclusion.tex
\section{Conclusions}

In this paper, we propose a view invariant 3D human pose estimation framework to advance the state-of-the-art. The VI-HC subnetwork which transforms the initial 3D poses to consistent views is designed to efficiently correct the 3D poses. A view invariant discriminator is introduced to impose high-level constraints over body configurations to improve the performance. Experimental results demonstrate that our proposed framework improves the performance significantly compared with the powerful baseline methods and is robust to different baseline methods. 